\documentclass{article}

\usepackage{neurips_2023}

\usepackage[utf8]{inputenc} % allow utf-8 input
\usepackage[T1]{fontenc}    % use 8-bit T1 fonts
\usepackage{hyperref}       % hyperlinks
\usepackage{url}            % simple URL typesetting
\usepackage{booktabs}       % professional-quality tables
\usepackage{amsfonts}       % blackboard math symbols
\usepackage{nicefrac}       % compact symbols for 1/2, etc.
\usepackage{microtype}      % microtypography
\usepackage{xcolor}         % colors

\usepackage{natbib}
\setcitestyle{square, comma, numbers,sort&compress, super}

\usepackage{graphicx}       % include graphics
\usepackage{subcaption}

\title{Super Resolution On Global Weather Forecasts}

\author{%
  Lawrence Zhang, Adam Yang, Rodz Andrie Amor, Bryan Zhang, Dhruv Rao \\
  University of Maryland \\
  Department of Computer Science
}

\begin{document}

\maketitle

% \begin{abstract}
  
% \end{abstract}

\section{Introduction}

Weather forecasting is a vitally important tool for tasks ranging from planning day to day activities to disaster response planning. However, modeling weather has proven to be challenging task due to its chaotic and unpredictable nature. Each variable, from temperature to precipitation to wind, all influence the path the environment will take. As a result, all models tend to rapidly lose accuracy as the temporal range of their forecasts increase. 

Classical forecasting methods use a myriad of physics-based, numerical, and stochastic techniques to predict the change in weather variables over time. However, such forecasts often require a very large amount of data and are extremely computationally expensive. Furthermore, as climate and global weather patterns change, classical models are substantially more difficult and time-consuming to update for changing environments.

Fortunately, with recent advances in deep learning and publicly available high quality weather datasets, deploying learning methods for estimating these complex systems has become feasible. The current state-of-the-art deep learning models have comparable accuracy to the industry standard numerical models and are becoming more ubiquitous in practice due to their adaptability. 

Recently, GraphCast demonstrated the ability to produce fast and accurate global weather predictions for up to 10 days into the future, using only very recent weather data as input. This has proven to be a monumental breakthrough for weather prediction research, especially with weather neural networks. However, while GraphCast has proven its usefulness for global weather forecasts, it does not have as much utility on a smaller scale because of its low resolution. This limits its value for the kind of day-to-day localized weather forecasting that the vast majority of people would find useful.

Our group seeks to improve upon existing deep learning based forecasting methods by increasing spatial resolutions of global weather predictions. Specifically, we are interested in performing super resolution (SR) on GraphCast temperature predictions by increasing the global precision from 1 degree of accuracy to 0.5 degrees, which is approximately 111km and 55km respectively.

\section{Motivation}
Global forecasts provide high level trends of climate throughout the entire planet. While such forecasts can model long range trends in weather, the results are too coarse for local weather applications. For instance, a model that produces weather forecast images at 1 degree spatial resolution will have pixels that cover over 100km. While much work has been done to create global forecasting models that have increased spatial and temporal accuracy, there has not been a high volume of work dedicated to modifying existing models to better meet the demands of localized forecasts. In this paper, we explore using deep learning methods to increase the spatial resolution of coarse global forecasts by increasing the precision of image forecasts produced by Graphcast %that has potential for general use for a variety of climate predictive systems.

This task can be described as performing SR and downscaling from low resolution to high resolution. Inspired by a paper on deep convolutional SR networks \cite{dong2015image}, we propose using a similar SR model that is specifically trained to artificially downscale forecasts at higher spatial resolution. Instead of training a larger forecasting model that come at the cost of energy and training time, lightweight SR models as a cheap alternative to extract weather features to create higher resolution predictions. This offers the benefit of the fast global weather forecasting with the higher granularity of larger models.

\section{Prior Work and State of the Art}
\subsection{GraphCast \cite{graphcast}}
GraphCast is the current standard for graph-based neural network global weather forecasting model. It is a common benchmark for weather forecasts because of its relatively low computation requirement and ability to produce a global medium-range 10-day forecast in under a minute on a Google Cloud Tensor Processing Unit. The model uses an autoregressive approach to make further date forecasts, utilizing its own previous six-hour forecast as the input by projecting variables onto a mesh and decoding learned features back onto a standard latitude-longitude grid. To evaluate the accuracy of the predictions, the researchers calculated the root mean squared error and the anomaly correlation coefficient to that of other models with the HRES dataset as the baseline. GraphCast demonstrated a statistically significant improvement in accuracy in all weather variables, and it subsequently became the benchmark standard machine learning weather prediction model. Using a frozen GraphCast, we are forwarding weather data to create forecasts on temperature, precipitation, and cloud coverage which will then be upscaled.

\subsection{Image Super-Resolution Using Deep Convolutional Networks \cite{dong2015image}}
This paper proposes a computationally lightweight but remarkably effective convolutional neural network that transforms a low resolution image to a high resolution image. The SRCNN consists of three major steps: patch extraction, non-linear mapping, and reconstruction. It learns the end-to-end mapping from low resolution inputs to high resolution outputs. This is especially useful for applications in deblurring or upscaling images. Finally, the SRCNN architecture performs well under various benchmarks as well as metrics like MSE and Peak Signal-to-Noise Ratio (PSNR). We use this architecture to perform the image SR from the global weather forecast data to a higher detail. However, this paper uses a 3 channel mapping to represent RGB whereas global weather data comes in 1 channel, which can represent a variable like temperature or cloud coverage. We have modified the architecture used in this paper to work with one channel input and output, as well as added a series of residual blocks and skip connections to improve its performance.

\subsection{Implicit Neural Representations with Periodic
Activation Functions \cite{sitzmann2020implicit}}
This paper proposes a new approach to generating Implicit Neural Representations (INR) by using periodic activation functions. INR approximate a continuous function from data that is represented discretely through using simple fully conected Multi-Layer Perceptrons as universal approximation functions. The goal is for the Multi-Layer Perceptron to learn a mapping between a discrete input to a continuous output.
Notably, this paper highlights the significant effectiveness of Multi-Layer Perceptrons as INRs when using a sine activation function (SIRENs) which is infinitely differentiable, with each derivative being a smooth function. Essentially, this means that SIRENs are good at modeling smooth surfaces and transitions. These INRs have direct applications within the field of SR due to their ability to map pixel coordinates directly to approximated values. 

\subsection{GAN based Super Resolution \cite{ledig2016gan}}
This paper proposes a super-resolution generative adversarial network (SRGAN) that utilizes a deep residual network with skip connections. Rather than using MSE loss, the researchers devised a perceptual loss centered around high-level feature maps and a discriminator that tries to ensure generated images are similar to input ones. The perceptual loss is a weighted linear combination of standard content loss from most other deep learning networks and adversarial loss, or loss calculated from probabilities of the binary classification the discriminator is performing. The generator architecture consists of residual blocks followed by 2 convolutional layers that extract features and produce higher-resolution images, and the discriminator is a series of convolutional layers ending with a sigmoid function to extract binary classification probabilities. For testing, in addition to analyzing the content loss over time, the researchers employed human subjects to score the quality of generated images from different models on a scale from 1 to 5. The researchers found that the model performs favorably on image datasets in comparison with other leading models of the time, such as ResNets.
% \subsection{Upscaled GraphCast \cite{oskarsson2023graphbased}}

%Replace this with a super image resolution 

\section{Data}
The base GraphCast forecasts numerous variables for a myriad atmospheric levels which are included in ERA5 data. We decided to select a few variables based on their relative importance for local areas. In particular, we selected, cloud coverage, and temperature 2 meters above sea level, as we believe that these features have the most impact on day-to-day life. Furthermore, each variable has its own challenges and advantages. We found that global cloud coverage had the best visualization and was the easiest to interpret visually, while temperature had the best qualities for testing upscaling due to its global patterns and relative spatio-temporal variance. 

The GraphCast model is trained on ERA5 model, which makes it particularly suited to work with that dataset. ERA5 is incredibly detailed, well documented, and readily available for any use, which enabled us to experiment with the dataset \cite{era5}. Fortunately, with the structure the ERA5 dataset takes and the information GraphCast can make inputs on, we were able to create a one-to-one mapping from GraphCast output to the corresponding ERA5 ground truth. 
%Another high quality data source that matched well with GraphCast is the MSWEP dataset. MSWEP is a modified version of the ERA5 dataset but contains information pooled from a variety of our data sources. However, during the development process, we found that many of the modifications made in MSWEP made the data source incompatible with our base frozen GraphCast. As a result, we decided to only utilize the pure ERA5 dataset for training and testing.

In our data pipeline, the first step we implemented collects a large number of inferences from the base GraphCast output. By doing so, we do not have to perform a full GraphCast inference every time we want to load input data into a data loader. We limited inferences to only output our relevant features in order to reduce the size of the dataset. Next, we made a function that uses the date labeled on the output file to match it with the corresponding ERA5 file. Finally, we implemented a data loader that maps the GraphCast output with the exact corresponding ERA5 data to use as the input and ground truth pair. 

\section{Methods}
% To train our models, we produced inferences from GraphCast over 12 hour intervals for 200 days and 2 images a day, totaling to 400 images. Additionally to supplement training we also used raw ERA5 forecast data across a 6 month and 1 year timeframe for temperature and cloud coverage respectively, with the time stamp for each image used to collect corresponding ERA5 ground truth images. The objective of the model then is to minimize the error when upscaling GraphCast image to ERA5 ground truth. We trained and compared various models
Instead of modifying the architecture of GraphCast to work with different resolutions, we decided to maintain a frozen GraphCast and use it as a black box for the input of our model. Then, we developed a SR model that works with the output of base GraphCast. This enables us to easily modify the structure of our network without having to deal with the larger, complex GraphCast architecture. In order to obtain a dataset to train our models, we produced inferences from GraphCast over 12 hour intervals for 200 days and 2 images a day, totaling to 400 images. These inferences were saved in a Google Drive folder to be later pulled and used in a data loader for training, validating, and testing. Next, we use the time stamp label for each image to map to the corresponding ERA5 images as the ground truth. The objective of the model is to minimize the error when upscaling GraphCast image to ERA5 ground truth. For the loss functions, we experimented with mean squared error, mean absolute error, edge error, and perception error. In the architecture fine-tuning part of the process, our group has experimented with a variety of models and loss functions. We experimented on UNet, SRCNN, GAN, and INR, with varying levels of success. The following subsections will go into further detail on the process and evaluation of the architectures we implemented.

\subsection{UNet}
The UNet architecture \cite{ronneberger2015unet} was previously developed for object segmentation tasks. The original paper that introduced it displayed use cases in biomedical image processing, but it was quickly adopted for a wide variety of other tasks. The high level architecture uses convolutional layers and pooling to produce a low level feature map of an image with decreased spatial dimension compared to the original image passed in. The model then up-samples the feature map input resolution back to the original spatial dimensions and uses a final softmax layer to produce segmentation results. For SR tasks, the output can be transformed to a regression objective where outputs model a continuous weather variable. 

We decided to first experiment with the UNet architecture because of its relative simplicity compared to other deep learning models and our previous familiarity with it. We hoped to establish it as a relatively straightforward base before exploring other, more complex models. However, outputs of the UNet model seemed to average the inputs and were unable to capture original image features. The output of the UNet simply was uninterpretable and showed no sign of capturing the patterns and upscaling the weather data. As a result, we started experimenting with other models listed below.

\subsection{SRGAN}
The next model we investigated was a SR generative adversarial network (SRGAN). The idea behind a GAN is to have two competing networks being trained at the same time, also known as adversarial training \cite{ledig2016gan}. The generator network generates new images by making modifications to the input data, whereas the discriminator network tries to predict whether the generated data came from the input dataset or not. Ideally, the goal is to have the generator producing data that the discriminator cannot tell apart from the actual input. This has immediate benefits in image generation tasks, where we can assess the quality of generated images by seeing if the discriminator can distinguish them from input images, as well as from the accuracy of the prediction. In the context of SR models, the generator would be identical to the SRCNN architecture discussed in the next section, but now the discriminator can be used to verify the quality of the images the SRCNN creates. This discriminator utilizes convolutional layers to extract features from the generated image before performing a binary classification using the sigmoid activation function to classify the image as real input or generated, first proposed by Ledig, et al. (2016)\cite{ledig2016gan}. If the discriminator cannot tell the difference between the two, the SRCNN generator component has produced a high-quality SR. However, we had recurring problems with implementing the GAN and ensuring that the model produced high quality and detailed images. In our case, the generator could not create any image complex enough to fool the discriminator, and the discriminator maintained a very low loss and high accuracy throughout the entire training process. With this in mind, we pivoted to experimenting with training the SRCNN with MSE error in the next section.

\subsection{SRCNN}
Finally, our best performing model is the SR convolutional neural network (SRCNN). This was a novel approach first devised by Dong, et al. (2014) \cite{dong2015image} that utilizes a relatively simple architecture of 3 main layers to perform image SR. After a bicubic interpolation to upscale the input image to the desired size resulting in a pixelated low resolution map, the model extracts feature maps using convolutional layers, nonlinearly maps these feature maps to another set of high-dimensional feature blocks, and then constructs the final higher-resolution image. Later papers built on this baseline by making the model deeper using more convolutional layers, and then a paper by Tong, et al. (2017) \cite{skip_connections} brought the idea of multiple skip connections into the fold to protect against vanishing gradients and overfitting. Work by Song, et al. (2019) \cite{song2019efficient} studied in more depth the idea of residual blocks, or networks where information could move from initial layers to end layers without any restriction through skip connections and then be combined with information that did pass through all the layers. 

For our implementation, we found that the base SRCNN model from Dong, et al. worked remarkably well, but was not complex enough to capture the smaller details and specific patterns. As a result, in order to increase the model complexity, we have added residual blocks and skip connections. The intent of the skip connections is to preserve the features in later layers by directly bypassing the layers. Similarly, the residual blocks work almost identically to the skip connections, but the residuals blocks also contain a convolution that is intended to learn the small distinctions between the input and output. By adding these two layers into the base SRCNN, we were able to have it produce images in more detail and capture smaller patterns.

\newpage
\section{Results}
To train our models, we used Mean Squared Error (MSE) loss to minimize the difference between outputs and ground truth and as an evaluation metric. Across the three tested models, SRCNN performed the best in matching the ground truth model according to our evaluation metric. 

\begin{figure}[h]
    \centering
    \includegraphics[width=.9\linewidth]{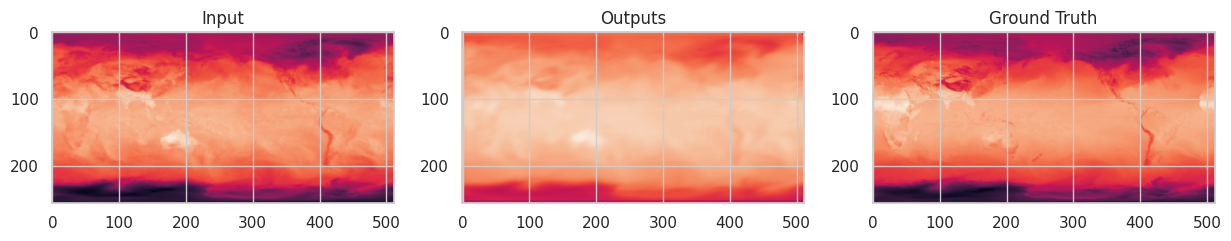}
    \label{fig:enter-label}
    \centering
    \includegraphics[width=.9\linewidth]{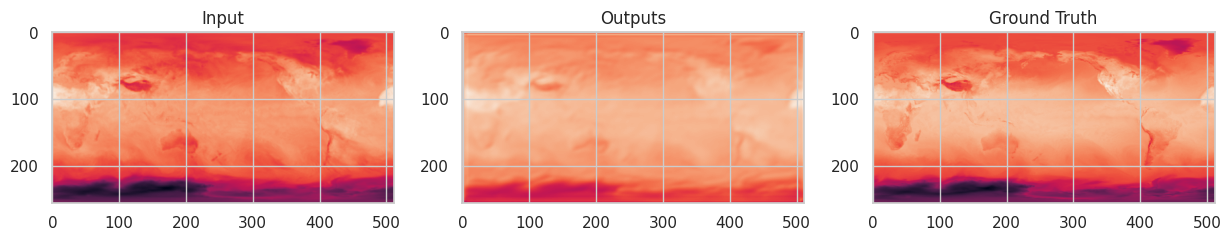}
    \label{fig:enter-label}
    \centering
    \includegraphics[width=.9\linewidth]{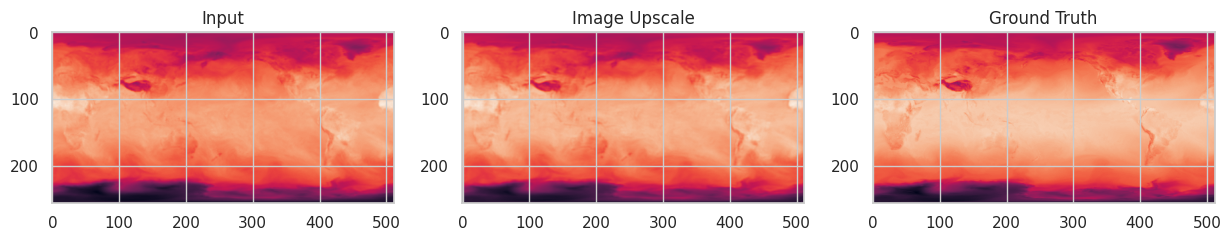}
    \caption{Results from UNet, SRGAN, and SRCNN Respectively}
    \label{fig:enter-label}
\end{figure}

Additionally, we were able to demonstrate from ERA5 low resolution data of cloud coverage to a higher resolution ERA5 cloud coverage. This demonstrates how the model is capable of downscaling any type of weather data and is not restricted to global temperature.

\begin{figure}[h!]
    \centering
    \includegraphics[width=.75\linewidth]{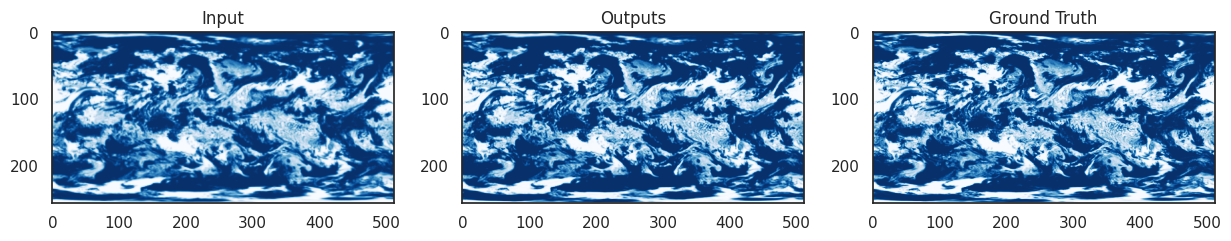}
    \caption{Global Cloud Coverage Upscale}
    \label{fig:enter-label}
\end{figure}

 \begin{figure}[h!]
     \centering
     \includegraphics[width=.75\linewidth]{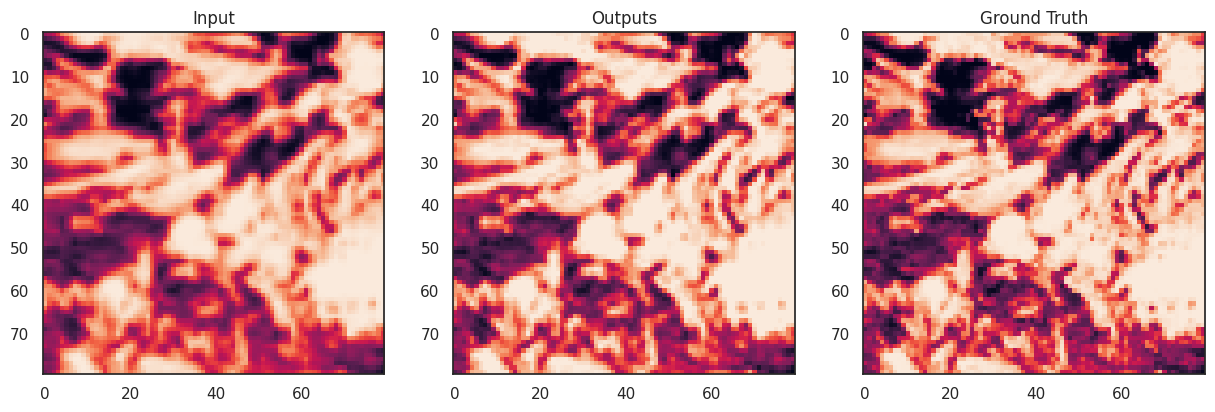}
     \caption{High Contrast Blowup of Image 80x80}
     \label{fig:enter-label}
\end{figure}

\begin{figure}[h!]
    \centering
    \includegraphics[width=.75\linewidth]{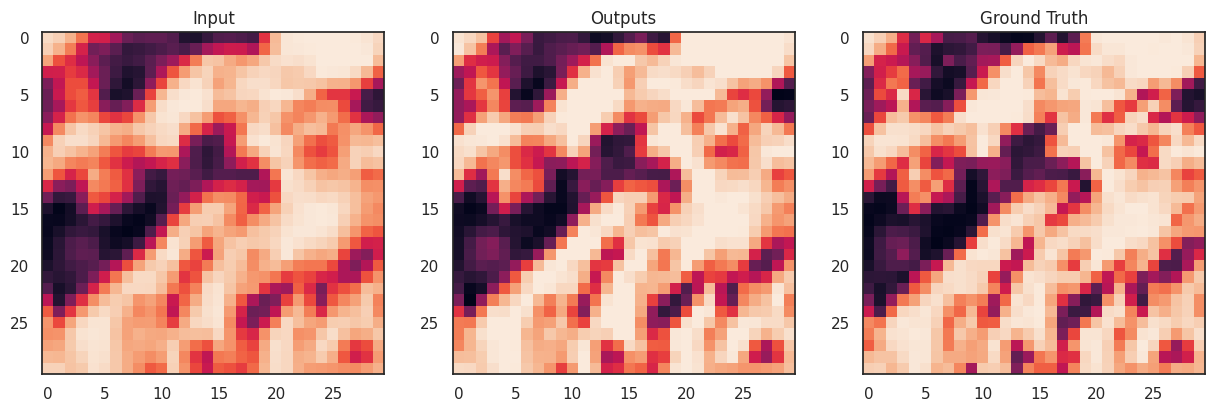}
    \caption{High Contrast Blowup of Image 30x30}
    \label{fig:enter-label}
\end{figure}
\newpage

\section{Super Resolution from INR with SIREN/WIRE}

An alternative method for approaching SR tasks that doesn't utilize traditional CNNs is by using INRs to approximate a continuous function from data that is represented discretely. This has direct applications to our problem of upscaling the weather data from ERA5 as these continuous functions are able to make images of any resolution by interpolating on the function. 

Our approach was to train multiple INRs using periodic and non-periodic activation functions. The use of periodic activation functions, such as sinusoidal and Gaussian, was explored by Sitzmann et al. \cite{sitzmann2020implicit} as an alternative to non-periodic activation functions, such as ReLU and Sigmoid, when designing an implicit representation. 

\subsection{Architecture}

\begin{figure}[h!]
    \centering
    \includegraphics[width=0.75\linewidth]{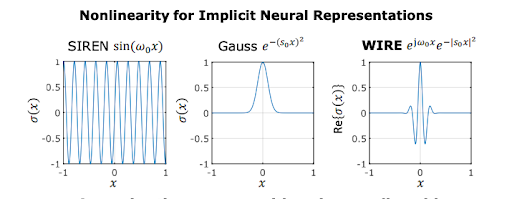}
    \caption{Non-Linear Activation Functions}
    \label{fig:enter-label}
\end{figure}

The underlying architecture of the INRs is the standard multi-layer perceptrons with 2 hidden layers, each hidden layer having 256 neurons. We trained 4 separate models using the same architecture but alternated between 4 different activations:
\begin{itemize}
    \setlength\itemsep{0.1em}
    \item ReLU (non-periodic)
    \item Gauss (periodic)
    \item Sinusoidal (periodic)
    \item Complex Gabor Wavelet (periodic) 
\end{itemize}

\subsection{Results}

After training each INR model for 2000 epochs, the models using periodic activation functions (Gauss, Siren, Wire) significantly outperformed the model using the ReLU activation when trying to upscale an image of cloud formations from 64 x 128 to 256 x 512. These results are consistent with previous experiments by Sitzmann et al.\cite{sitzmann2020implicit}

% \begin{figure}[htbp]
%   \caption{Super Resolution Evaluation Metrics}
%   \begin{subfigure}[b]{0.3\textwidth}
%     \includegraphics[width=\textwidth]{images/inr_mse.png}
%     \caption{Mean Squared Error}
%     \label{fig:image1}
%   \end{subfigure}
%   \begin{subfigure}[b]{0.3\textwidth}
%     \includegraphics[width=\textwidth]{images/inr_psnr.png}
%     \caption{Peak Signal To Noise Ratio}
%     \label{fig:image1}
%   \end{subfigure}
%     \begin{subfigure}[b]{0.3\textwidth}
%     \includegraphics[width=\textwidth]{images/inr_ssim.png}
%     \caption{Structural Similarity Index}
%     \label{fig:image1}
%   \end{subfigure}
% \end{figure}

When comparing across the different periodic activation functions using PSNR and SSIM as the evaluation metrics, the results are not consistent with Saragadam et al. \cite{saragadam2023wire}. Unlike their experiment, the SIREN model performed slightly better than WIRE here for our SR task, but this may be due to the different attributes of weather images and the training images in the researcher's dataset. When comparing the results of the INRs and SRCNN model, we found that the INRs had a lower MSE score across all the periodic activations. 

% \begin{figure}[h!]
%     \centering
%     \includegraphics[width=.9\linewidth]{images/inr_cloud_formations.png}
%     \caption{Results from INR Models}
%     \label{fig:enter-label}
% \end{figure}

\begin{figure}[h!]
  \centering
  \begin{subfigure}[b]{.9\textwidth}
    \centering
    \includegraphics[width=.9\linewidth]{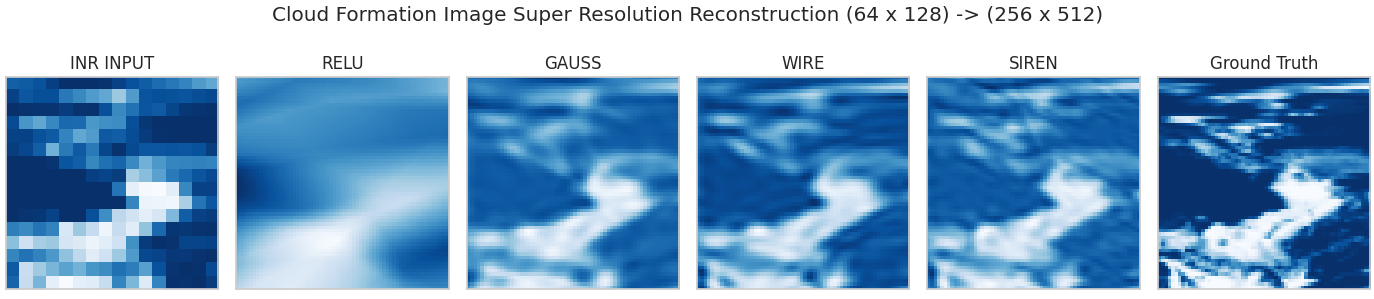}
    \caption{Results from INR Models}
    \label{fig:enter-label}
  \end{subfigure}
  \begin{subfigure}[b]{.75\textwidth}
  \centering
    \includegraphics[width=\textwidth]{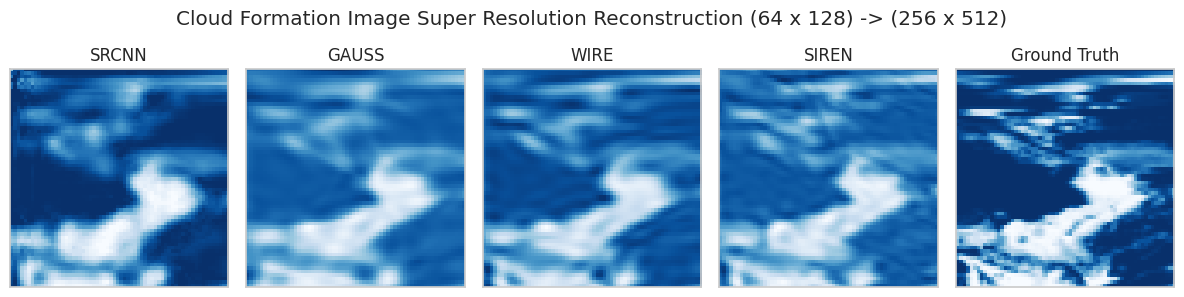}
    \caption{SRCNN and Periodic INR results}
    \label{fig:image1}
  \end{subfigure}
\end{figure}

% \begin{figure}[htbp]
%   \centering
%   \caption{SRCNN and INRs}
%   \begin{subfigure}[b]{.75\textwidth}
%     \includegraphics[width=\textwidth]{images/inr_srcnn.png}
%     \caption{SRCNN and Periodic INR results}
%     \label{fig:image1}
%   \end{subfigure}
%   \begin{subfigure}[b]{0.5\textwidth}
%     \includegraphics[width=\textwidth]{images/inr_srcnn_mse.png}
%     \caption{INR and SRCNN Mean Squared Error}
%     \label{fig:image1}
%   \end{subfigure}
% \end{figure}

% \begin{figure}[h]
%     \centering  
%     \includegraphics[width=0.5\linewidth]{images/inr_srcnn_mse.png}
%     \caption{INR and SRCNN Mean Squared Error}
%     \label{fig:enter-label}
% \end{figure}

% \begin{figure}[h]
%     \centering
%     \includegraphics[width=.75\linewidth]{images/inr_srcnn.png}
%     \caption{SRCNN and Periodic INR results}
%     \label{fig:enter-label}
% \end{figure}

The last experiment we ran with the INR was on 2m temperature (Temperature 2 meters above the surface level). We trained on the same SR task going from 64 x 128 to 256 x 512 and found that the WIRE and SIREN activations performed just as well as the cloud formation upscaling.

\begin{figure}[h]
    \centering
    \includegraphics[width=.75\linewidth]{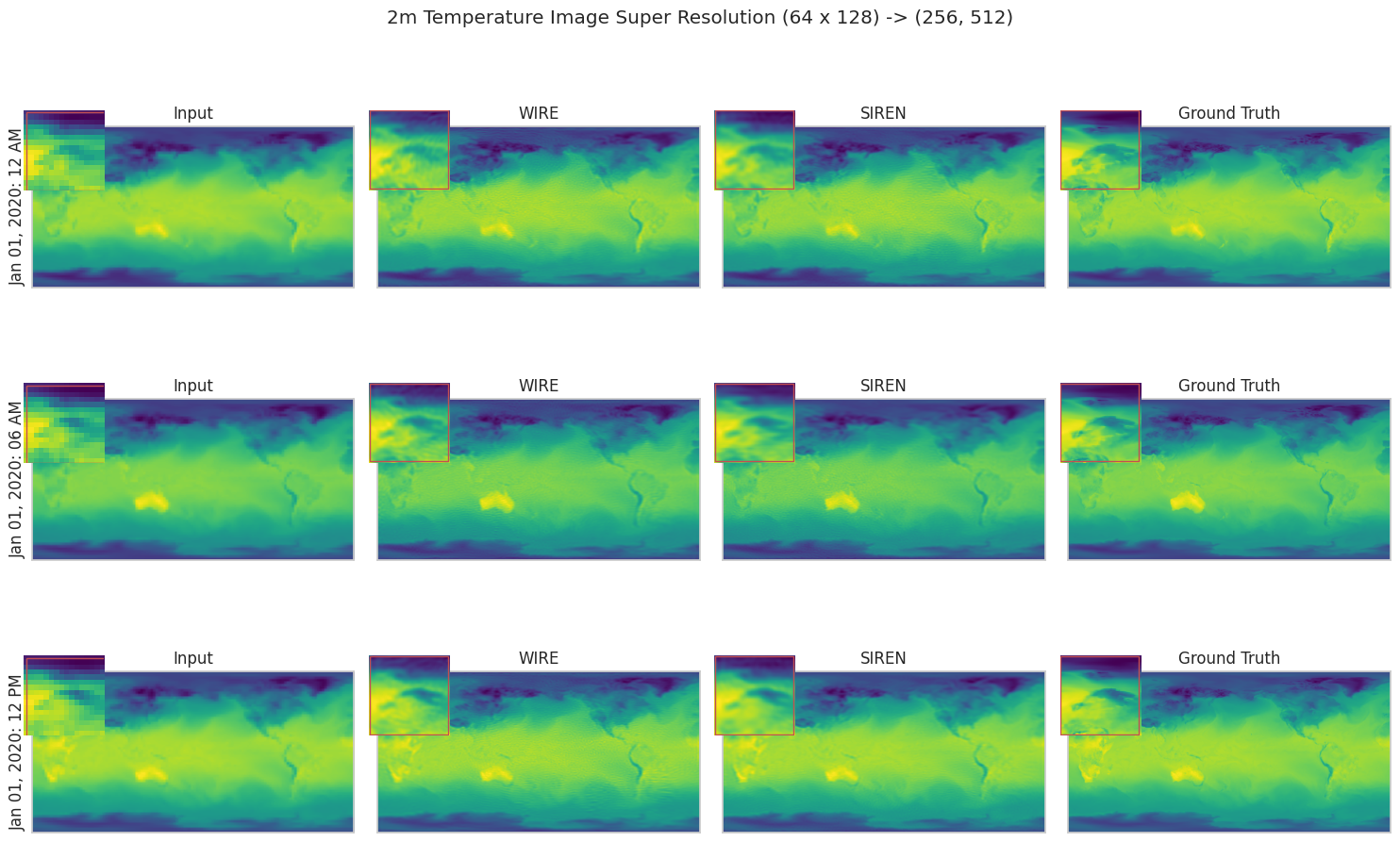}
    \caption{Surface Temperature SR}
    \label{fig:enter-label}
\end{figure}

\section{Future work} 
\subsection{Multi Modal Data}
While our models were able to produce reasonable estimations for a single variable, future work can seek SR methods that integrate multi-modal inputs. Image to image tasks are typical for SR objectives, but the high dimensionality of weather data offers a chance for more robust models trained to measure associations between various measurements. For our data pipeline, the architectures were only trained and tested on single channel images. However, a model capable of processing multimodal data will be able to recognize the correlations between variables and produce more fine-grain results.  

\subsection{Localized Models}
While models such as GraphCast provide global forecasts, more lightweight and local models may be trained on limited areas. Using local models may prove more accurate for smaller regions while costing a fraction of running inference for global forecasts. Furthermore, local models may capture specific and nuanced patterns only seen by in an area that the global models are not suited to recognize. By restricting the boundary to a specific area and increasing the resolution, it will be substantially more informative for local residents. For very small areas, a precision of 0.1 degrees, or roughly 11 kilometers, and even lower will certainly be significantly more useful.

\subsection{Autoregressive SR}
An additional method of SR is to approach the task in an autoregressive style training scheme. Theoretically, this method would progressively denoise inputs and produce high fidelity outputs over a number of iterations. We demonstrate this with our SRCNN model, but found that training repeated iterations would be computationally expensive. In addition, the model is prone to introducing unwanted artifacts and remove accurate predictions over numerous iterations. However, at a controlled amount, the model does potentially show increased precision and is especially useful in non-noisy regions especially 

\begin{figure}[h]
    \centering  
    \includegraphics[width=0.75\textwidth]{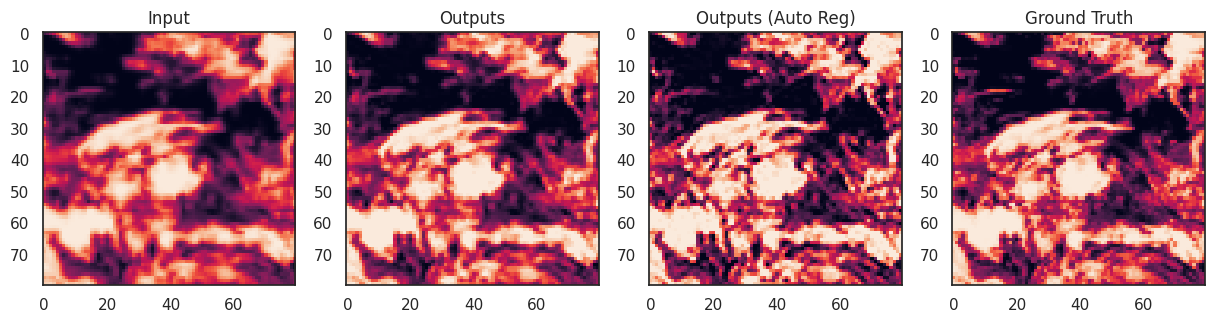}
    \caption{Autoregressive SRCNN}
    \label{fig:enter-label}
\end{figure}

\newpage

\bibliographystyle{plainnat}  
\bibliography{references}

\end{document}